\documentclass[10pt,twocolumn,letterpaper]{article}

\usepackage{wacv}
\usepackage{times}
\usepackage{epsfig}
\usepackage{graphicx}
\usepackage{amsmath}
\usepackage{amssymb}

\usepackage{multirow, booktabs, tabularx}
\usepackage{amsfonts}
\usepackage{siunitx}
\usepackage{soul}
\usepackage{subcaption}
\usepackage[nameinlink]{cleveref}
\crefname{figure}{Fig.}{Figs.}
\crefname{table}{Tab.}{Tabs.}
\crefname{section}{Sec.}{Secs.}
\crefname{subsection}{Sec.}{Secs.}

\usepackage{fixmath}
\renewcommand{\vec}[1]{\mathbold{#1}}

\newcommand{\txt}{\mathbold{r}}
\newcommand{\txtFeat}{\mathbold{p}}

\newcommand{\txtReal}{\mathbold{r^+}}
\newcommand{\txtRealFeat}{\mathbold{p}^+}

\newcommand{\txtWrong}{\mathbold{r}^-}
\newcommand{\txtWrongFeat}{\mathbold{p}^-}

\newcommand{\img}{\mathbold{v}}
\newcommand{\imgFeat}{\mathbold{q}}

\newcommand{\imgReal}{\mathbold{v^+}}
\newcommand{\imgRealFeat}{\mathbold{q}^+}

\newcommand{\imgWrong}{\mathbold{v^-}}
\newcommand{\imgWrongFeat}{\mathbold{q}^-}

\newcommand{\imgFake}{\mathbold{\tilde{v}^+}}
\newcommand{\imgFakeFeat}{\mathbold{\tilde{q}}^+}

\DeclareMathOperator{\R}{\mathbb{R}}
\DeclareMathOperator{\TxtEnc}{F_p}
\DeclareMathOperator{\ImgEnc}{F_q}
\DeclareMathOperator{\Ncal}{\mathcal{N}}
\DeclareMathOperator{\Lcal}{\mathcal{L}}
\DeclareMathOperator*{\softmax}{softmax}

\newcommand{\best}[1] {\num[math-rm=\mathbf]{#1}}

\usepackage{xspace}
\newcommand{\FoodSpace}{\texttt{FoodSpace}\xspace}

\wacvfinalcopy 

\def\wacvPaperID{96} 

\ifwacvfinal\fi
\begin{document}

\title{CookGAN: Meal Image Synthesis from Ingredients}

\author{Fangda Han \\
Rutgers University, \\
Piscataway, NJ, USA\\
{\tt\small fh199@cs.rutgers.edu}
\and
Ricardo Guerrero \\
Samsung AI Center, Cambridge, UK\\
{\tt\small r.guerrero@samsung.com}
\and
Vladimir Pavlovic \\
Rutgers University,\\
Piscataway, NJ, USA\\
{\tt\small vladimir@cs.rutgers.edu}
}

\maketitle
\ifwacvfinal\thispagestyle{empty}\fi

\begin{abstract}
   In this work we propose a new computational framework, based on generative deep models, for synthesis of photo-realistic food meal images from textual list of its ingredients. Previous works on synthesis of images from text typically rely on pre-trained text models to extract text features, followed by generative neural networks (GAN) aimed to generate realistic images conditioned on the text features. These works mainly focus on generating spatially compact and well-defined categories of objects, such as birds or flowers, but meal images are significantly more complex, consisting of multiple ingredients whose appearance and spatial qualities are further modified by cooking methods. To generate real-like meal images from ingredients, we propose Cook Generative Adversarial Networks (CookGAN), CookGAN first builds an attention-based ingredients-image association model, which is then used to condition a generative neural network tasked with synthesizing meal images. Furthermore, a cycle-consistent constraint is added to further improve image quality and control appearance. Experiments show our model is able to generate meal images corresponding to the ingredients.
\end{abstract}

\section{Introduction}
\label{sec:intro}

Computational food analysis (CFA) has become a pivotal area for the computer vision community due to its real-world implications for nutritional health~\cite{carvalho2018cross, salvador2017learning, horita2018food, chen2018deep, ahn2011flavor, teng2012recipe, martinel2018wide, aguilar2018grab, DBLP:journals/corr/abs-1808-07202}. 
For instance, being able to extract food information, including ingredients and calories, from a meal image could help us monitor our daily nutrient intake and manage our diet. 
In addition to food intake logging, CFA can also be crucial for learning and assessing the functional similarity of ingredients, meal preference forecasting, and computational meal preparation and planning \cite{teng2012recipe, helmy2015health}.
The advancement of CFA depends on developing better frameworks that aim at extracting food-related information from different domains including text descriptions (\eg recipe title, ingredients, instructions) and meal images, as well as exploring the relationships between different domains in order to better understand food related properties.

This paper concentrates on generating meal images from a set of specific ingredients. Although image generation from text is popular in the computer vision community~\cite{reed2016generative, zhang2017stackgan, zhang2017stackgan++}, similar work on generating photo-realistic meal images has so far failed to materialize due to the complex factors associated to meal images, these factors include appearance diversity, dependency on the cooking method, variations in preparation style, visual dissolution, etc. 
As a consequence, the generative meal model has to infer these key pieces of information implicitly.

In this work, we propose Cook Generative Adversarial Networks (CookGAN), a model to generate a photo-realistic meal image conditioned on a list of ingredients (we will use `ingredient list' or `ingredients' interchangeably in this paper). The efficacy of the model is analyzed by modifying the visible ingredients. 
The main contributions are: 1) Combining attention-based recipe association model~\cite{chen2018deep} and StackGAN~\cite{zhang2017stackgan++} to generate meal images from ingredients. 2) Adding a cycle-consistency constraint to further improve image quality and control the appearance of the image by changing ingredients.

\begin{figure*}[ht]
\begin{center}
\includegraphics[width=\textwidth]{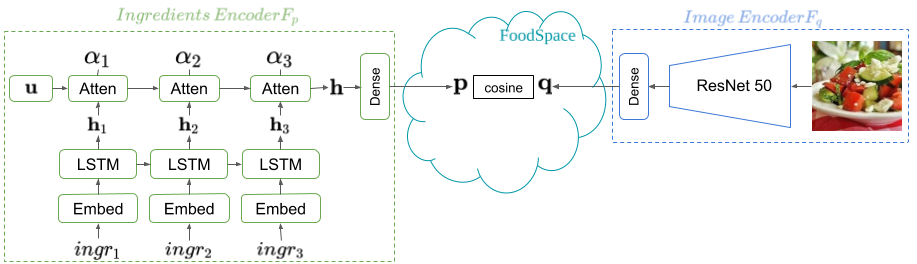}
\end{center}
\caption{The framework of the attention-based cross-modal association model.}
\label{fig:framework_association}
\end{figure*}
\section{Related Work}
\paragraph{Generative neural networks (GAN)} GAN is a popular type of generative model for image synthesis \cite{goodfellow2014generative}. 
It learns to model the distribution of real images via a combination of two networks, one that generates images from a random input vector and another that attempts to discriminate between real and generated images.

\paragraph{Conditional GAN} Work on generating images conditioned on a deterministic label by directly concatenating the label with the input was proposed by~\cite{mirza2014conditional} and by adding the label information at a certain layer's output in~\cite{odena2017conditional, miyato2018cgans}. 
Another line of work conditions the generation process with text information.
\cite{reed2016generative} uses a pre-trained model to extract text features and concatenate them with the input random vector, in order to generate text-based images.
\cite{zhang2017stackgan++} extends this concept by stacking three GAN to generate images at different resolutions.
These works are conditioned on short textual descriptions of the image and rely on recurrent neural networks (RNN) to extract text features. RNNs treat words sequentially and with the same importance.
However, in the sparse list of ingredients of a meal, not all ingredients occupy equally important roles in image appearance.
Inspired by \cite{chen2018deep}, we combine attention mechanism with bi-directional LSTM (a commonly used RNN model) to learn the importance of each ingredient. The attention mechanism helps locate key ingredients in an unsupervised way.

\paragraph{Meal Image Generation} Most prior work for image generation from text implicitly assume the visual categories are well-structured singular objects, consistent in appearance (\eg birds, flowers, or faces). Meal images, on the other hand, have more variable appearance when conditioned on ingredients.
\cite{wang2019learning} and \cite{zhu2019r2gan} use generative neural networks to generate meal images as a constraint to improve cross-modal recipe retrieval, however, they only generate low-resolution (\eg $128\times 128$) images, and furthermore, because the synthesized image is only used to regularize the retrieval model, image quality is not well evaluated. 
\cite{papadopoulos2019make} uses GAN to generate pizza images given step-by-step procedures, however, their model is only tested with pizza with a pre-defined list of procedures and ingredients. Compared with them, we aim at generating meal images with various food types and ingredients.

~\cite{wang2019food} and~\cite{el2019gilt} are more closely related to our work, however, we include a cycle-consistency regularizer to minimize the semantic discrepancy between fake and real images.
The guiding intuition is that if the generated image is of high quality and captures the ingredients correctly, it should extract similar feature as that from the real image. 
Experiment shows this regularizer improves image quality both qualitatively and quantitatively.


\section{Methodology}


To generate a meal image from an ingredient list, we first train an attention-based association model to find a shared latent space between ingredient list and image, then use the latent representation of the ingredient list to train a GAN to synthesize the meal image conditioned on the list.

\subsection{Attention-based Association Model}
\label{subsec:association_model}

In order to extract ingredient features, we train an attention-based cross-modal association model \cite{chen2018deep} to match an ingredient list and its corresponding image in a joint latent space, denoted the \FoodSpace.
During training, the model takes a triplet as input, which includes the recipe ingredient list, its corresponding image, and an image from another recipe, $(\txtReal, \imgReal, \imgWrong)$, respectively. Using two separate neural networks, one for the ingredient list $\TxtEnc$ and another for images $\ImgEnc$, the triplet is embedded in the \FoodSpace with coordinates $(\txtRealFeat, \imgRealFeat, \imgWrongFeat)$. The networks are trained to maximize the association in \FoodSpace between positive pair $(\txtRealFeat,\imgRealFeat)$, and at the same time minimizing the association between negative pair $(\txtRealFeat,\imgWrongFeat)$.

Formally, with the ingredients encoder $\txtFeat = \TxtEnc(\txt)$ and image encoder $\imgFeat = \ImgEnc(\img)$, the training is a maximization of the following objective function, 
\begin{align}
\begin{split}
    & V(\TxtEnc, \ImgEnc) = \\ 
    & \mathbb{E}_{ \hat{p}(\txtReal,\imgReal), \hat{p}(\imgWrong) } \min\left( \left[ d{\left[\txtRealFeat, \imgRealFeat\right]} - d{\left[\txtRealFeat, \imgWrongFeat\right]} - \epsilon \right], 0 \right) + \\
    & \mathbb{E}_{ \hat{p}(\txtReal,\imgReal), \hat{p}(\txtWrong) } \min\left( \left[ d{\left[\txtRealFeat, \imgRealFeat\right]} - d{\left[\txtWrongFeat, \imgRealFeat\right]} - \epsilon \right], 0 \right),
\end{split}
\end{align}
where $\cos{\left[\txtFeat, \imgFeat\right]} = \txtFeat^\intercal\imgFeat / \sqrt{ (\txtFeat^\intercal\txtFeat) (\imgFeat^\intercal\imgFeat)}$ is the cosine similarity in \FoodSpace and $\hat{p}$ denotes the corresponding empirical densities on the training set. We combine the cosine similarity of the positive pair and that of the negative pair together, and we add a margin $\epsilon$ to make the model focus on those pairs that are not correctly embedded. We empirically set $\epsilon$ to $0.3$ by cross-validation.
\cref{fig:framework_association} shows a diagram of the attention-based association model. The details of ingredients encoder $\TxtEnc$ and image encoder $\ImgEnc$ are explained below.

\textbf{Ingredients encoder} $\TxtEnc$ takes the recipe's ingredients as input and outputs their feature representation in \FoodSpace. The goal is to find the embedding that reflect dependencies between ingredients, which could facilitate implicit associations even when some ingredients are not visually observable. 
For this purpose, the model first embeds the one-hot vector of each ingredient into a low-dimension vector ($ingr_i \in \R^{300}$) using a word2vec model~\cite{mikolov2013efficient}, treating the vectors as a sequence input of a bi-directional LSTM\footnote{Hence, we assume a chain graph can approximate arbitrary ingredient dependencies within a recipe.}.
Instead of using the output of the last layer as the output of the LSTM, each hidden state $\vec{h}_i \in \R^{300}$ is used as the feature of the corresponding ingredient.

As not all ingredients play equally important roles in image appearance, we apply attention mechanism to model the contribution of each ingredient.
During training, the model learns a shared contextual vector $\vec{u} \in \R^{300}$ of the same dimension as the hidden state, and $\vec{u}$ is then used to assess the attention of each ingredient,
\begin{align}
    \left\{\alpha_i\right\} = \softmax\left\{ \vec{u}^T \cdot \vec{h}_i \right\}, \;i\in[1,N],
\end{align}
where $N$ is the number of ingredients in the recipe. The attention-based output of LSTM is a weighted summation of all hidden states, $\vec{h} = \sum_{i=1}^{N} \alpha_i \vec{h}_i$. The contextual vector $\vec{u}$ is optimized as a parameter during training and fixed during testing. Our intuition is $\vec{u}$ can attend on certain ingredients that appear in a specific ingredient list by learning from the training data.
Finally, $\vec{h}$ is projected to \FoodSpace to yield the ingredients feature $\txtFeat \in \R^{1024}$ . 

\textbf{Image encoder} $\ImgEnc$ takes a meal image as input and outputs a feature representing the image in \FoodSpace. Resnet50 \cite{he2016deep} pre-trained on ImageNet~\cite{girshick2014rich} is applied as the base model for feature extraction.
In order to get a more meaningful feature of the image, we follow \cite{chen2018deep} and finetune the network on UPMC-Food-101 \cite{wang2015recipe}, we use the activation after the average pooling ($\R^{2048}$) and project it to \FoodSpace to get $\imgFeat \in \R^{1024}$.

\subsection{Generative Meal Image Network}
\begin{figure*}[ht]
\begin{center}
\includegraphics[width=\textwidth]{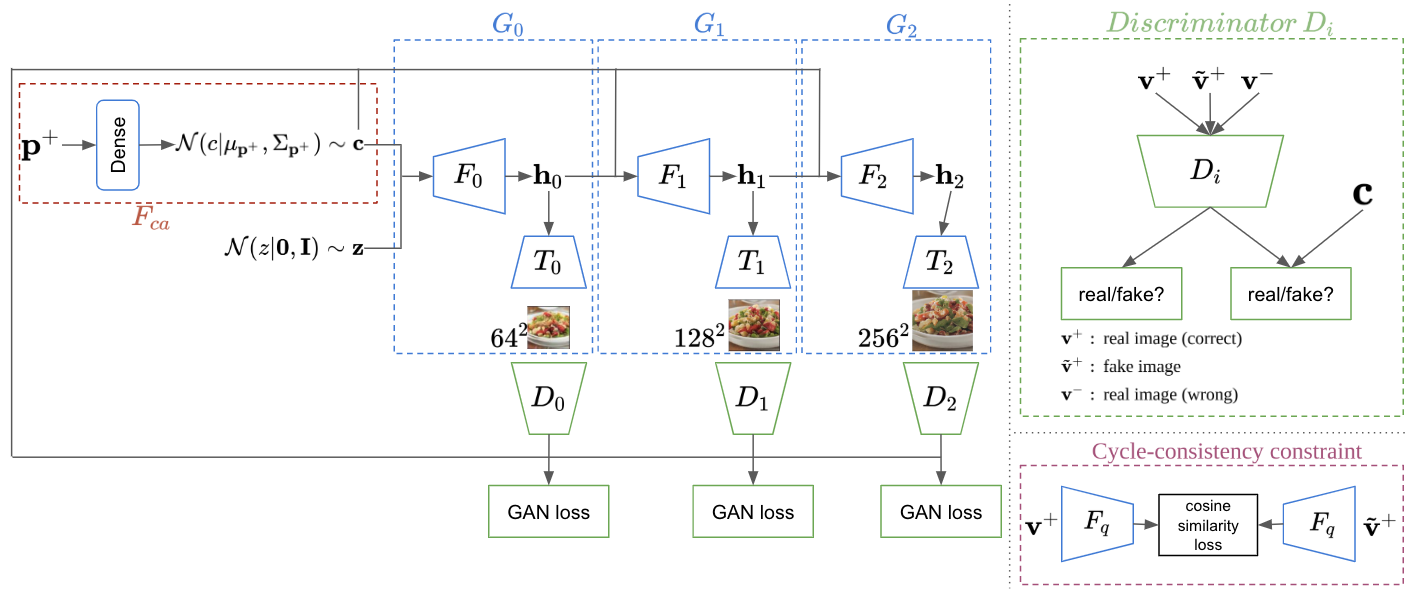}
\end{center}
\caption{Framework of the generative model. $G_0$, $G_1$, $G_2$ represent the branches in generator. $D_0$, $D_1$, $D_2$ represent the discriminators for images of low, medium and high resolution. $F_q$ is the image encoder trained in the association model.}
\label{fig:generator}
\end{figure*}

Generative meal image network takes the ingredient list as input and generates the corresponding meal image.
The base model StackGAN-v2~\cite{zhang2017stackgan++} contains three branches stacked together. Each branch is responsible for generating image at a specific scale and each branch has its own discriminator which is responsible for distinguish the image at that scale.
The framework is shown in \cref{fig:generator}.

\textbf{Generator:} The ingredients $\txtReal$ are first encoded using the pre-trained $\TxtEnc$ (fixed during training StackGAN-v2) to obtain text feature $\txtRealFeat$.
Subsequently, $\txtRealFeat$ is forwarded through a conditional augmentation network $F_{ca}$ to estimate the distribution $p(\vec{c}|\txtRealFeat)$ of the ingredient appearance factor $\vec{c}$, modeled as the Gaussian distribution
\begin{align}
    \left(\mu_{\txtRealFeat}, \Sigma_{\txtRealFeat}\right) &= F_{ca}(\txtRealFeat), \\
    \vec{c} \sim p(\vec{c}|\txtRealFeat) &= \Ncal(\mu_{\txtRealFeat}, \Sigma_{\txtRealFeat}),
\end{align} 
where $\mu_{\txtRealFeat}$ and $\Sigma_{\txtRealFeat}$ are the mean and the covariance given the ingredients encoding $\txtRealFeat$ in \FoodSpace. 
This sampling process introduces noise to $\txtRealFeat$,  making the model robust to small perturbations in \FoodSpace.
Variational regularization~\cite{kingma2013auto} is applied during training to make $p(c|\txtRealFeat)$ close to the standard Gaussian distribution,
\begin{align}
    \Lcal_{ca} = D_{KL} \left[ \Ncal(\mu_{\txtRealFeat}, \Sigma_{\txtRealFeat}) || \Ncal(\vec{0}, \vec{I})  \right].
\end{align}
Subsequently, $\vec{c}$ is augmented with Gaussian noise $\vec{z} \sim \Ncal(\vec{0}, \vec{I})$ to generate the latent feature $\vec{h}_0 = F_0(\vec{z}, \vec{c})$ for the first branch and the low-resolution image $\imgFake_0  = T_0(\vec{h}_0)$, where $F_0$ and $T_0$ are modeled by neural networks. Similarly, the medium and high resolution images are generated by utilizing the hidden feature of the previous branches, $\vec{h}_1 = F_1(\vec{h}_0, \vec{c}), \; \imgFake_1 = T_1(\vec{h}_1)$ and $\vec{h}_2 = F_2(\vec{h}_1, \vec{c}), \; \imgFake_2 = T_2(\vec{h}_2)$. Overall, the generator contains three branches, each responsible for generating the image at a specific scale, $G_0 = \{F_0, T_0 \} ,\; G_1 = \{ F_1, T_1 \} ,\; G_2 = \{ F_2, T_2 \}$. Optimization of the generator will be described after introducing the discriminators.

\textbf{Discriminator:} Each discriminator's task is three-fold: (1) Classify real, `correctly-paired' $\imgReal$ with ingredient appearance factor $\vec{c}$ as real; (2) Classify real, `wrongly-paired' $\imgWrong$ with $\vec{c}$ as fake; and (3) Classify generated image $\imgFake$ with $\vec{c}$ as fake. 
Formally, we seek to minimize the cross-entropy loss
\begin{align}\small
\label{eq:loss_d_cond}
\begin{split}
    \Lcal^{cond}_{i} = 
    & -\mathbb{E}_{\imgReal \sim p_{d_i}} [\log D_i(\imgReal, \vec{c})] \\
    & + \mathbb{E}_{\imgWrong \sim p_{d_i}} [\log D_i(\imgWrong, \vec{c})] \\
    & +\mathbb{E}_{\imgFake \sim p_{G_i}} [\log D_i(\imgFake, \vec{c})],
\end{split}
\end{align}
where $p_{d_i}$, $p_{G_i}$, $G_i$ and $D_i$ correspond to the real image distribution, fake image distribution, generator branch, and the discriminator at the $i^{th}$ scale.
To further improve the quality of the generated image, we also minimize the unconditional image distribution as
\begin{align}\small
\label{eq:loss_d_uncond}
\begin{split}
    \Lcal^{uncond}_{i} = 
    & -\mathbb{E}_{\imgReal \sim p_{d_i}} [\log D_i(\imgReal)] \\ 
    & - \mathbb{E}_{\imgWrong \sim p_{d_i}} [\log D_i(\imgWrong)] \\
    & + \mathbb{E}_{\imgFake \sim p_{G_i}} [\log D_i(\imgFake)]
\end{split}
\end{align}

\textbf{Losses:} During training, the generator and discriminators are optimized alternatively by maximizing and minimizing \eqref{eq:loss_d_cond} and \eqref{eq:loss_d_uncond} respectively. All generator branches are trained jointly as are the three discriminators, with final losses
\begin{align}
    \label{eq:l_g}
    \Lcal_G &= \sum_{i=0}^2 \left\{ \Lcal^{cond}_{i} + \lambda_{uncond} \Lcal^{uncond}_{i}\right\} + \lambda_{ca}\Lcal_{ca} \\
    \label{eq:l_d}
    \Lcal_D &= \sum_{i=0}^2 \left\{ \Lcal^{cond}_{i} + \lambda_{uncond} \Lcal^{uncond}_{i}\right\},
\end{align}
where $\lambda_{uncond}$ is the weight of the unconditional loss and $\lambda_{ca}$ the weight of the conditional augmentation loss.
We empirically set $\lambda_{uncond}=0.5$ and $\lambda_{ca}=0.02$ by cross-validation.

\subsection{Cycle-consistency constraint}
\label{sec:cycle}
A correctly-generated meal image should "contain" the ingredients it is conditioned on. Thus,
a cycle-consistency term is introduced to keep the fake image contextually similar, in terms of ingredients, to the corresponding real image in \FoodSpace.

Specifically, for a real image $\imgReal$ with \FoodSpace coordinate  $\imgRealFeat$ and the corresponding generated $\imgFake$ with $\imgFakeFeat$, the cycle-consistency regularization aims at maximizing the cosine similarity at different scales, $\Lcal_{C_i} = \cos{\left[\imgRealFeat, \imgFakeFeat\right]}$. Note that the images in different resolutions need to be rescaled for the input of the image encoder.
The final generator loss in \eqref{eq:l_g} now becomes 
\begin{align}
\begin{split}
    \label{eq:l_g_with_cycle}
    \Lcal_G = 
    &\sum_{i=0}^2 \left\{ \Lcal^{cond}_{i} + \lambda_{uncond} \Lcal^{uncond}_{i} - \lambda_{cycle} \Lcal_{C_i} \right\} \\
    & + \lambda_{ca}\Lcal_{ca},
\end{split}
\end{align}
where $\lambda_{cycle}$ is the weight of the cycle-consistency term, cross-validated to $\lambda_{cycle}=1.0$.

\section{Experiments}
\begin{table*}[ht]
\sisetup{round-mode=places,round-precision=3}
    \centering
    \begin{tabular}{llrrrrrrrr}
        \toprule
        && \multicolumn{4}{c}{im2recipe} & \multicolumn{4}{c}{recipe2im} \\ \cmidrule{3-10}
        && MedR$\downarrow$ & R@1$\uparrow$ & R@5$\uparrow$ & R@10$\uparrow$ & MedR$\downarrow$ & R@1$\uparrow$ & R@5$\uparrow$ & R@10$\uparrow$\\
        \midrule
        \multirow{2}{*}{1K} 
        & attention \cite{chen2018deep} &-&-&-&-&-&-&-&- \\ \cmidrule{2-10}
        & ours w/o attn & \best{5.4} & \num{0.2294} & \best{0.5102} & \best{0.6207} & \best{5.7360} & \num{0.2298} & \best{0.5015} & \num{0.6096} \\ \cmidrule{2-10}
        & ours w/ attn & \num{5.5} & \best{0.2342} & \num{0.5025} & \num{0.6182} & \num{5.7500} & \best{0.2303} & \num{0.4910} & \best{0.6152} \\
        \midrule
        \multirow{2}{*}{5K} 
        & attention \cite{chen2018deep} & \num{71.0} & \num{0.045} & \num{0.135} & \num{0.202} & \num{70.1} & \num{0.042} & \num{0.133} & \num{0.202} \\ \cmidrule{2-10}
        & ours w/o attn & \best{24.0} & \best{0.1053} & \num{0.2602} & \num{0.3598} & \num{25.3} & \num{0.0939} & \best{0.2607} & \best{0.3578} \\ \cmidrule{2-10}
        & ours w/ attn & \best{24.0} & \num{0.0986} & \best{0.2653} & \best{0.3642} & \best{25.1000} & \best{0.0965} & \num{0.2585} & \num{0.3574} \\
        \midrule
        \multirow{2}{*}{10K} 
        & attention \cite{chen2018deep} & - & - & - & - & - & - & - & - \\ \cmidrule{2-10}
        & ours w/o attn & \best{47.5} & \num{0.0648} & \num{0.1834} & \best{0.2700} & \num{48.5} & \num{0.0607} & \best{0.1892} & \best{0.2715} \\ \cmidrule{2-10}
        & ours w/ attn & \num{47.7} & \best{0.0653} & \best{0.1846} & \num{0.2670} & \best{48.3000} & \best{0.0614} & \num{0.1777} & \num{0.2613} \\
        \bottomrule
    \end{tabular}
    \caption{Comparison with the baseline~\cite{chen2018deep} for using image as query to retrieve recipe and vice versa. `w/o attn' means without attention, `w/ attn' means with attention. `$\downarrow$' means the lower the better, `$\uparrow$' means the higher the better, `-' stands for score not reported in~\cite{chen2018deep}.}
    \label{tab:comparison_retrieval}
\end{table*}

\textbf{Dataset} Data used in this work was taken from Recipe1M \cite{salvador2017learning}. This dataset contains $\sim$1M recipes with titles, instructions, ingredients, and images.
We focus on a subset of \num{402760} recipes with at least one image, containing no more than 20 ingredients or instructions, and no less than one ingredient and instruction.
Data is split into 70\% train, 15\% validation and 15\% test sets, using at most \num{5} images from each recipe. 

Recipe1M contains $\sim$\num{16}k unique ingredients, we reduce this number by focusing on the 4k most frequent ones. 
This list is further reduced by first merging the ingredients with the same name after a stemming operation and semi-automatically fusing other ingredients. 
The later is achieved using a word2vec model trained on Recipe1M, where the ingredients are fused if they are close together in their embedding space and a human annotator accepts the proposed merger.
Finally, we obtain a list of \num{1989} canonical ingredients, covering more than \SI{95}{\%} of all recipes in the dataset.

\textbf{Implementation Details} Specific network structures follow those in \cite{chen2018deep} for the association model\footnote{Here we use LSTM instead of GRU because they have similar performance as stated in their paper.} and \cite{zhang2017stackgan++} for the generator.

\subsection{Effect of Canonical Ingredients}
\label{sec:association_model_evaluation_canonical_ingrs_effect}

\begin{figure}[ht]
\begin{center}
\includegraphics[width=8.0cm]{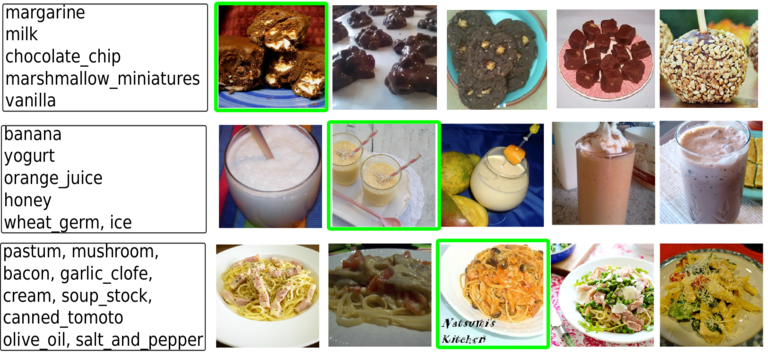}
\end{center}
\caption{Sample results of using ingredients as query to retrieve images on a 1K dataset. Left: query ingredients. Right: top 5 retrieved images (sorted). Corresponding image is indicated by the green box.}
\label{fig:retrieval_examples}
\end{figure}

To evaluate the effect of the proposed canonical ingredients, we compare with the attention-based model~\cite{chen2018deep}. Given a query in one modality, the goal is to retrieve the paired point in the other modality by comparing their similarities in \FoodSpace.
The association model is trained on four Tesla K80 for 16 hours (25 epochs) until convergence. 

\textbf{Metrics.} We applied the same metrics as~\cite{chen2018deep}, including the median retrieval rank (MedR) and the recall at top K (R@K). 
MedR is computed as the median rank of the true positive over all queries, a lower MedR $\ge$\num{1.0} suggests better performance. 
R@K computes the fraction of true positives recalled among the top-K retrieved candidates, it is a value between \num{0} to \num{100} with the higher score indicating better performance.

\textbf{Results.} In \cref{tab:comparison_retrieval}, we report the scores of the baseline model~\cite{chen2018deep} and that of the same model with our canonical ingredients (with and without attention). The performance is greatly improved on 5K samples, which clearly shows the advantage of using our canonical ingredients instead of the raw ingredients data.
\cref{fig:retrieval_examples} illustrates the top 5 retrieved images using the ingredients as the query.
Although the correct images do not always appear in the first position, the retrieved images largely belong to the same food type, suggesting commonality in ingredients.

\subsection{Effect of Attention}
\label{sec:association_model_evaluation_attention_effect}
\begin{figure*}[ht]
\includegraphics[width=\textwidth]{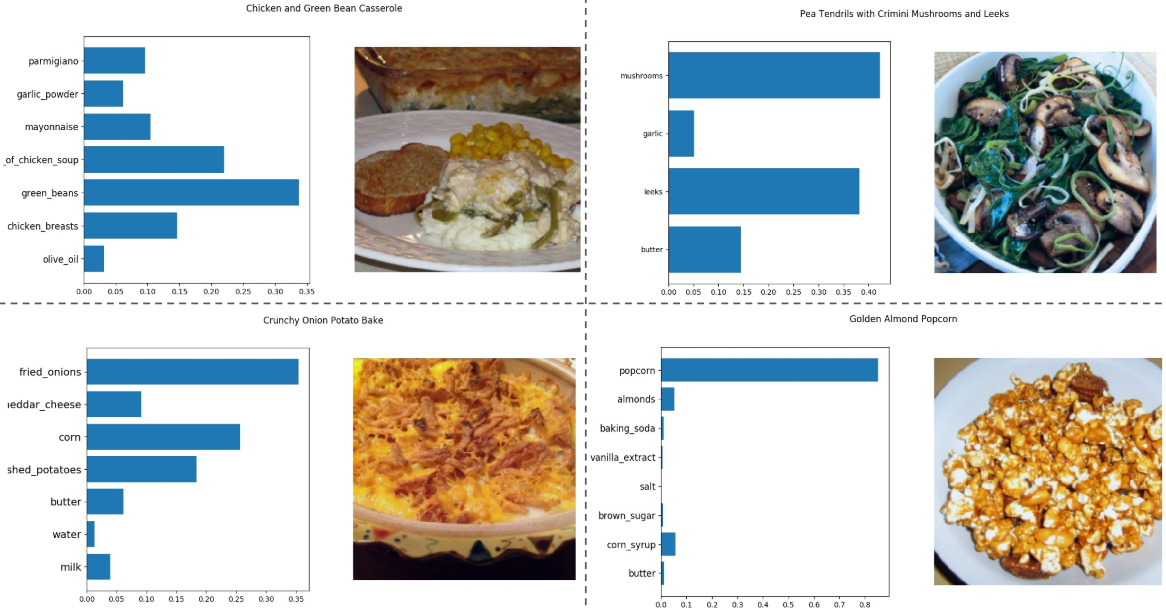}
\caption{Attention of the ingredients.}
\label{fig:attention}
\end{figure*}

\begin{table*}[ht]
\sisetup{round-mode=places,round-precision=2, separate-uncertainty}

\begin{tabular}[t]{cc}
\begin{subtable}[t]{0.48\textwidth}
    \centering
    \begin{tabular}[t]{clrrr}
        \toprule
         & & salad & cookie & muffin \\
        \midrule
        \multirow{3}{*}{IS $\uparrow$}  & StackGAN-v2 & \num{3.0652} & \num[math-rm=\mathbf]{4.6992}  & \num{2.6020} \\ \cmidrule{2-5}
                                        & ours w/ CI & \best{3.4573} & \num{2.8172} & \best{2.9368} \\ \cmidrule{2-5}
                                        & ours w/o CI & \num{3.2856} & \num{3.5326} & \num{2.7888} \\ \cmidrule{2-5}
                                        & real & \num{5.1205} & \num{5.6989} & \num{4.1968} \\
        \midrule
        \multirow{2}{*}{FID $\downarrow$} & StackGAN-v2 & \best{55.4263} & \num{106.1381} & \num{104.7341} \\ \cmidrule{2-5}
                                            & ours w/ CI & \num{78.7922} & \best{87.1371} & \num{81.1320} \\ \cmidrule{2-5}
                                            & ours w/o CI & \num{62.6347} & \num{89.3295} & \best{80.2234} \\
        \bottomrule
    \end{tabular}
    \caption{Inception score (IS) and Frechet inception distance (FID).}
    \label{tab:is_and_fid}
\end{subtable}
&
\begin{subtable}[t]{0.48\textwidth}
    \centering
    \begin{tabular}[t]{lrrr}
        \toprule
         & salad & cookie & muffin \\
         \midrule
        random      & \num{450.0}   & \num{450.0}   & \num{450.0}   \\
        StackGAN-v2  & \best{58.4000}  & \num{194.45}  & \num{217.5000}  \\
        ours w/ CI    & \num{66.1500}   & \best{103.300}   & \best{211.0000}   \\
        ours w/o CI    & \num{82.4200}   & \num{125.2300}   & \num{232.3000}   \\
        real        & \num{12.15}   & \num{47.35}   & \num{65.0}    \\
        \bottomrule
    \end{tabular}
    \caption{Median rank comparison.}
    \label{tab:comparison_retrieval_synthesis}
\end{subtable}
\end{tabular}
    \caption{Performance analysis: (a) Comparison of StackGAN-v2 and our model on different subsets by inception scores (IS) and Frechet inception distance (FID). (b) Comparison of median rank (MedR) by using synthesized images to retrieve recipes in subsets. We choose $900$ as the retrieval range to adhere to the maximum number of recipes among test-sets for salad, cookie and muffin. `w/ CI' means with canonical ingredients, `w/o CI' means without canonical ingredients.}
\end{table*}

To evaluate the effect of the attention mechanism mentioned in \cref{subsec:association_model}, we report the performance of our model for retrieval with or without attention. Interestingly, our model with attention does not achieve better performance. 
This is somewhat counter-intuitive since it can be seen in \cref{fig:attention} that the model with attention tends to focus on visually important ingredients. 
%
%
For example, in top-left recipe, the model attends on green beans and chicken soup; in top-right recipe, the model attends on mushroom and leeks.
It should be noted that the model does not simply attend on ingredients that appears more frequently in the dataset (\eg olive\_oil, water, butter) but learns to focus on the ingredients that are more visible for the recipe.
We suspect the reason that attention mechanism does not improve to the performance scores is that the RNN model learns the importance of each ingredient implicitly.
Nevertheless, the attention mechanism can exist as an unsupervised method to locate important ingredients for a recipe.

\subsection{Meal Image Generation}

\begin{figure*}[ht]
    \includegraphics[width=\textwidth]{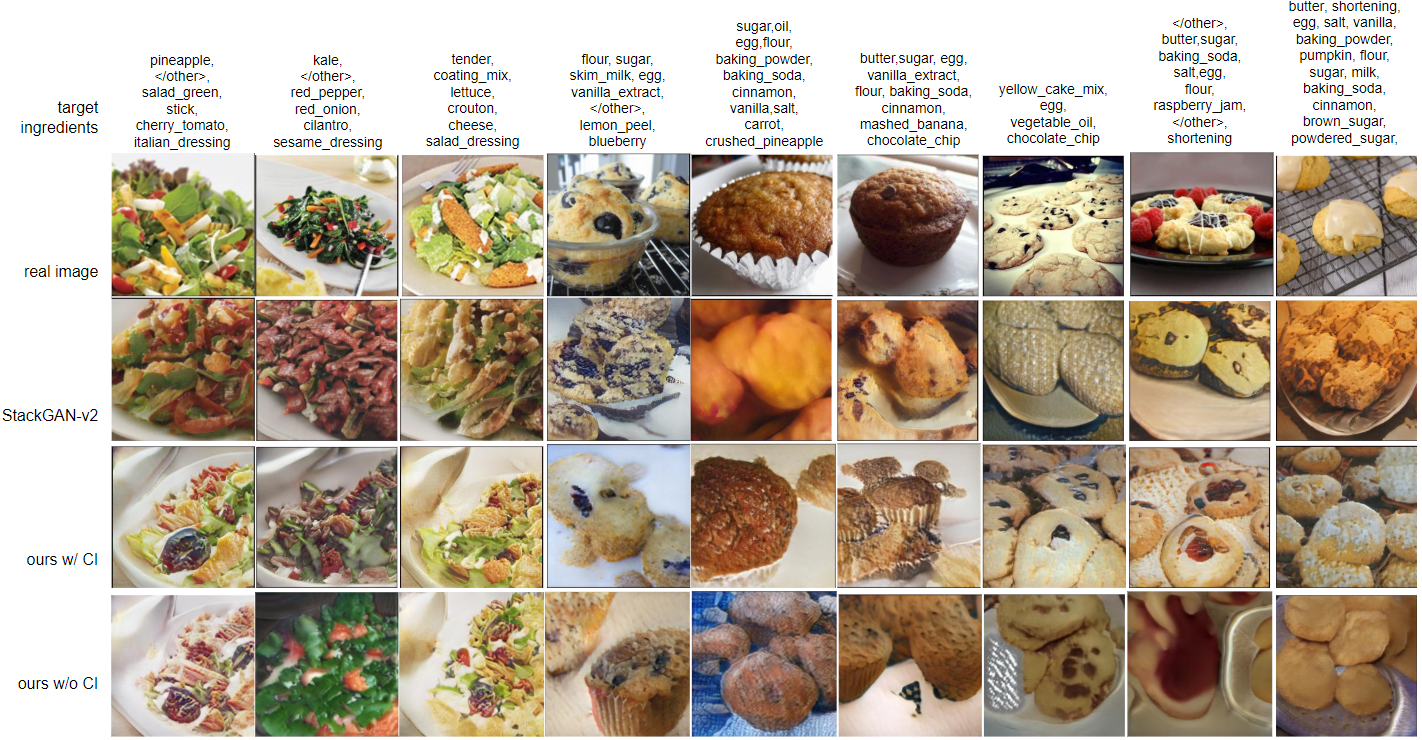}
    \caption{Example results by StackGAN-v2~\cite{zhang2017stackgan++} and our model conditioned on target ingredients, the real images are also shown for reference.}
    \label{fig:example_results}
\end{figure*}

\begin{figure}[ht]
\begin{center}
\includegraphics[width=0.40\textwidth]{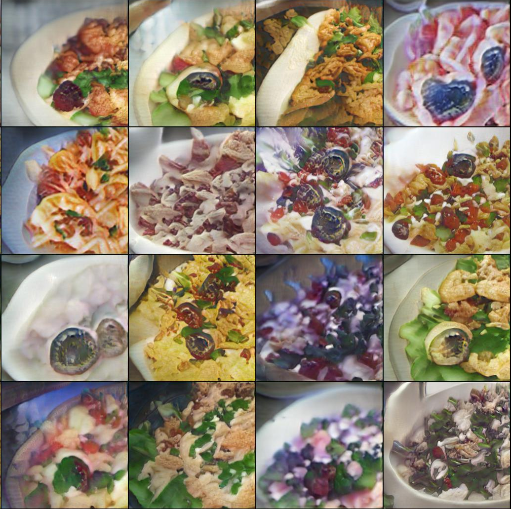}
\end{center}
\caption{Example results from different ingredients $\mathbf{c}$ with same random vector $\mathbf{z}$ in the salad subset.}
\label{fig:different_ingredients_same_noises_salad}
\end{figure}

\begin{figure}[ht]
\begin{center}
\includegraphics[width=0.45\textwidth]{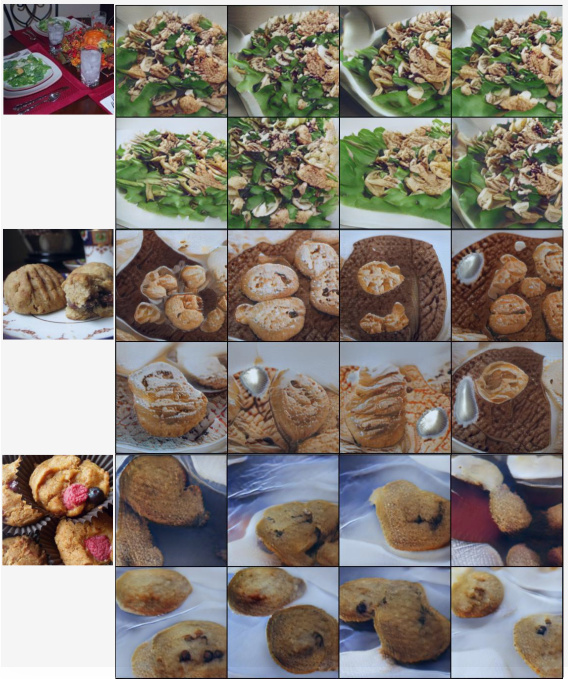}
\end{center}
\caption{Example results from same ingredients with different random vectors. \num{8} synthesized images are shown for each real image (top-left).}
\label{fig:same_ingredients_different_noises_wacv}
\end{figure}

\begin{figure}[ht]
\begin{center}
\includegraphics[width=7.6cm]{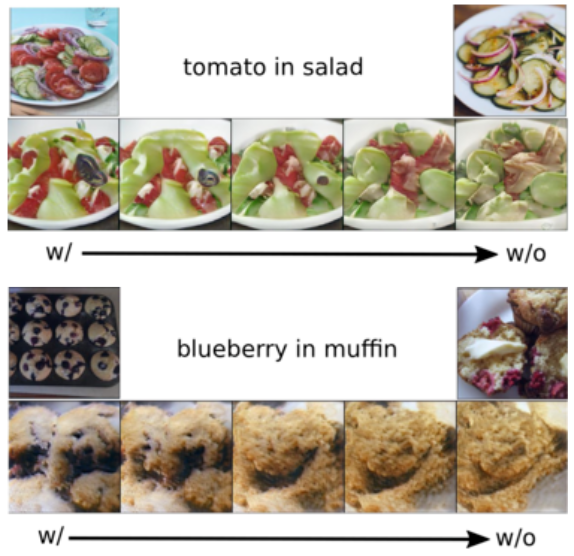}
\end{center}
\caption{Example results of synthesized images from the linear interpolations in \FoodSpace between two recipes (with and without target ingredient). Target ingredient on the left is tomato and the model is trained with salad subset; target ingredient on the right is blueberry and the model is trained with muffin subset. The interpolation points from left to right are $\frac{with}{without} = \left\{ \frac{4}{0}, \frac{3}{1}. \frac{2}{2}, \frac{1}{3}, \frac{0}{4} \right\}$}
\label{fig:linear_interpolations}
\end{figure}

We present the results of synthesizing meal image given an ingredient list. 
To mitigate the diversity caused by different preparation methods, we focus on narrow meal categories where the cutting and cooking methods are largely consistent within each category. 
In the following experiments, we only train on specific types of food within three commonly-seen categories: salad, cookie, and muffin. 
Images from these categories usually contain key ingredients that are easily recognized, which can be used to verify the model's ability to manipulate meal image by changing those ingredients. 
The number of samples in train/test dataset are $17209/3784$ (salad), $9546/2063$ (cookie) and $4312/900$ (muffin).

\textbf{Metrics.} Evaluating the performance of synthesized images is generally a challenging task due to the high complexity of images.
We choose Inception Score (IS) \cite{salimans2016improved} and Frechet Inception Distance (FID) \cite{heusel2017gans} as our quantitative evaluation metrics.



\textbf{Results.} We computed IS and FID on \num{900} samples randomly generated on the test-set for each category, which is the maximum number of recipes among test-sets for salad, cookie and muffin.
The IS of real images are also computed as a baseline.
\cref{tab:is_and_fid} shows the results obtained on different categories. 
We compare with StackGAN-v2~\cite{zhang2017stackgan++}, one of the state-of-the-art GAN model for text-based image synthesis. \textbf{ours w/o CI} uses the original ingredients and the proposed cycle-consistency constraint, while \textbf{ours w/ CI} uses the canonical ingredients and the proposed cycle-consistency constraint. We observe the model achieves better IS and FID on most subsets by using cycle-consistency constraint.
However, using canonical ingredients does not always lead to better scores for the generative model. We argue that image quality is more related to the design of the generative model while the canonical ingredients help more on the conditioning on the text.

To evaluate the conditioning on the text, we investigate the median rank (MedR) by using synthesized images as the query to retrieve recipes with the association model in \cref{subsec:association_model}.
\cref{tab:comparison_retrieval_synthesis} suggests using cycle-consistency constraint outperforms the baseline StackGAN-v2~\cite{zhang2017stackgan++} on most subsets, indicating the utility of the ingredient cycle-consistency. We also observe that applying canonical ingredients always leads to better MedR which demonstrates the effectiveness of our canonical-ingredients-based text embedding model. Still, the generated images remain apart from the real images in their retrieval ability, affirming the extreme difficulty of the photo-realistic meal image synthesis task.

\cref{fig:example_results} shows examples generated from different subsets. Within each category, the generated images capture the main ingredients for different recipes. Compared with StackGAN-v2~\cite{zhang2017stackgan++}, the images generated using cycle-consistency constraint usually have more clear ingredients appearance and looks more photo-realistic.




\subsection{Components Analysis}
Our generative model in~\cref{fig:generator} has two inputs, an ingredients feature $\mathbf{c}$ and a random vector $\mathbf{z}$. In this section we analyze the different roles played by these two components.

\cref{fig:different_ingredients_same_noises_salad} shows examples generated from different ingredients with the same random vector $\vec{z}$ in the salad subset.
the generated images contains different ingredients for different recipes while sharing a similar view point. This demonstrates the model's ability to synthesize meal images conditioned on ingredient features $\vec{c}$ while keeping nuisance factors fixed through vector $\vec{z}$.

\cref{fig:same_ingredients_different_noises_wacv} further demonstrates the different roles of ingredients appearance $\vec{c}$ and random vector $\vec{z}$ by showing examples generated from same ingredients with different random vectors. 
The synthesized images have different view points, but still all appear to share the same ingredients.

To demonstrate the ability to synthesize meal images corresponding to specific key ingredient, we choose a target ingredient and show the synthesized images of linear interpolations between a pair of ingredient lists $r_i$ and $r_j$ (in the feature space), in which $r_i$ contains the target ingredient and $r_j$ is without it, but shares at least \SI{70}{\%} of remaining ingredients in common with $r_i$\footnote{The reason for choosing the partial overlap is because very few recipes differ in exactly one key ingredient.}. One can observe that the model gradually removes the target ingredient during the interpolation-based removal process, as seen in \cref{fig:linear_interpolations}.


\section{Conclusion}
In this paper, we develop a model for generating photo-realistic meal images based on sets of ingredients. 
We integrate the attention-based recipe association model with StackGAN-v2, aiming for the association model to yield the ingredients feature close to the real meal image in \FoodSpace, with StackGAN-v2 attempting to reproduce this image class from the \FoodSpace encoding. To improve the quality of generated images, we reuse the image encoder in the association model and design an ingredient cycle-consistency regularization term in the shared space. Finally, we demonstrate that processing the ingredients into a canonical vocabulary is a critical key step in the synthesis process.
Experimental results demonstrate that our model is able to synthesize natural-looking meal images corresponding to desired ingredients, both visually and quantitatively, through retrieval metrics.
In the future, we aim at adding additional information including recipe instructions and titles to further contextualize the factors such as the meal preparation, as well as combining the amount of each ingredient to synthesize images with arbitrary ingredients quantities.

{\small
\bibliographystyle{ieee}
\bibliography{egbib}
}

\end{document}